\documentclass{article} 
\usepackage{iclr2015,times}
\usepackage{hyperref}
\usepackage{url}
\usepackage{bm}
\usepackage{amsmath}

\title{Generative Modeling of Hidden Functional Brain Networks}

\author{
Shaurabh Nandy\\
School of Behavioral and Brain Sciences\\
University of Texas at Dallas\\
Richardson, Texas 75080, USA \\
\texttt{shaurabh@utdallas.edu}\\
\And
Richard M. Golden\\
Human Language Technology Research Institute \\
University of Texas at Dallas\\
BBS, GR4.1, 800 West Campbell Road\\
Richardson, Texas, 75080, USA\\
\texttt{golden@utdallas.edu}\\
}

\iclrconference 

\begin{document}

\maketitle

\begin{abstract}
Resting-state functional connectivity fMRI data is a derivative of the unobservable neuronal functional network structure of the human brain. This data is subject to multiple sources of noise such as thermal noise, system noise, and physiological noise. Commonly used methods to infer the latent network structure, such as thresholding methods, make the implicit assumption that weak links are not as important as strong links, and that links are conditionally independent. However, such assumptions provide an incomplete description of the biology. Additionally, despite a core set of observations about functional networks such as small-worldness, modularity, exponentially truncated degree distributions, and presence of various types of hubs, very little is known about the computational principles which can give rise to these observations. This paper presents a Hidden Markov Random Field framework for the purpose of representing, estimating, and evaluating latent neuronal functional relationships using fMRI data. The main theoretical contributions of this paper are summarized as follows. 1) Provides a method to learn a more robust representation of the latent network structure. 2) Provides a method for testing multiple competing hypotheses of functional connectivity principles. 3) Provides a method which makes an explicit distinction between blood flow fMRI variables and unobservable neuronal activity variables. 4) Provides a method to model the conditional dependence structure between functional links, thereby assuming a more biologically plausible architecture.

\end{abstract}

\section{Introduction}
Functional magnetic resonance imaging (fMRI) is a very important neuroimaging technology because it allows us to non-invasively measure whole brain activity in humans. The crucial idea in fMRI is the concept of the blood oxygen level dependent (BOLD) signal. Neuronal activity requires oxygen which is provided by oxygenated hemoglobin, and in the aftermath of neuronal activity, the area of activity is overcompensated by a fresh supply of oxygenated blood. This overcompensation, the mechanisms of which are largely unknown, leads to a difference in ratio between oxygenated and de-oxygenated hemoglobin in the close vicinity of the brain region that was active. Using properties of magnetic resonance, it is then possible to pick up these differences and image brain activity. The signal which captures this difference is called the BOLD signal. With a spatial resolution in the order of millimeters (approx. 3mm) and temporal resolution in the order of seconds (approx. 1-6 seconds), fMRI complements other technologies such as electroencephalography (EEG) and positron emission tomography (PET). For example, EEG has excellent temporal resolution (in the milliseconds), but suffers in the domain of spatial resolution. Alternatively, PET provides better spatial resolution than EEG but its temporal resolution can run into the minutes (approx. 90 seconds to 30 minutes depending on the type of tracer used).

As fMRI has established itself as a dominant human brain imaging technology, there has also been a steady increase in the number of fMRI data analysis techniques in the toolkit of the practitioner. For example, compared to the initial days where the general linear model (GLM) dominated, it is now very common to see applications of more advanced techniques such as Structural Equation Modeling (Schlosser et al., 2003; Kim et al., 2007), Granger Causality (Roebroeck et al., 2005; Hamilton et al., 2011), Independent Component Analysis (Beckmann et al., 1997; Stone et al., 2002), Markov Chains (Allen et al., 2012) and Support Vector Machines (De Martino et al., 2008; LaConte et al., 2005). The increased repertoire of tools has greatly increased the questions one can ask using fMRI data. For example, Nishimoto et al. (2011) were able to show that it is possible to reconstruct dynamic images, and Cowen et al. (2014) demonstrated that it is possible to reconstruct face images, just from the fMRI activity evoked by such stimuli. These are just two out of many impressive innovations in the analysis of fMRI data.\nocite{schlosser2003altered} \nocite{kim2007unified} \nocite{roebroeck2005mapping} \nocite{hamilton2011investigating} \nocite{beckmann2005tensorial} \nocite{stone2002spatiotemporal} \nocite{allen2012tracking} \nocite{laconte2005support} \nocite{de2008combining} \nocite{nishimoto2011reconstructing} \nocite{cowen2014neural}

The earliest paradigm of fMRI was concentrated on task based studies, where different experimental tasks were correlated with spatial patterns of brain activation, thereby suggesting which regions of the brain are involved in the information processing demands of various task domains such as memory, attention, decision making etc. Since then the use of fMRI has expanded, and functional connectivity information can now be collected using a non-task paradigm called resting state. In this paradigm, data is collected from subjects who lie inside a fMRI scanner and are asked to fixate on a cross-hair or keep their eyes closed. There is no information processing demand. Subjects lie in the scanner for a couple of minutes to an hour depending on what the researcher considers an appropriate amount of time (see Smith et al. (2011) \nocite{smith2011network} for discussion on length of resting state scans). The utility of the resting state paradigm lies in the fact that regions of the brain that share function are correlated in the low frequency components of the BOLD signal (Cordes et al., 2001). \nocite{cordes2001frequencies} This crucial observation allows us to further our understanding of functional brain organization by looking at functional organization holistically, rather than in the isolation of particular task domains. The raw data, i.e., the time series of the low frequency component (\textless  0.1 Hz) of the resting state BOLD signal, is first pre-processed to remove artifacts, and then an appropriate measure of interdependency (such as Pearson's correlation coefficient) is used to quantify the functional connectivity between the various regions of interest. 

Hidden Markov Random Field's (HMRF's), which are latent variable probabilistic graphical models, are another class of models which have previously been applied to brain imaging data analysis. For example, HMRF's have been used for segmentation of MR images into tissue subclasses (Zhang et al., 2001; Esteban et al., 2014;) and functional parcellation of the cortex (Honnorat et al., 2015; Ryali et al., 2013). In this paper, we develop a novel HMRF framework for two purposes. Acknowledging that the BOLD signal is a proxy for neuronal activity and that resting state data is a noisy representation of the latent neuronal functional network, the first goal of this framework is to provide a method to learn a more robust representation of the latent structure. The second goal, which is intimately connected to the first goal, is to provide a method for testing multiple competing theories of functional connectivity. HMRF's have never before been applied to tackle the above mentioned goals. To motivate the importance of these goals, relevant background is first provided in the next section. In light of the background information, the goals of this paper are reiterated in greater detail in the section following the background, before finally diving into the mathematical development. \nocite{zhang2001segmentation} \nocite{esteban2014mbis} \nocite{honnorat2015grasp} \nocite{ryali2013parcellation}

\section{Background}


\subsection{Brain Connectivity}
The study of brain connectivity is usually divided into three major classes: structural connectivity, functional connectivity and effective connectivity. Structural connectivity refers to anatomical connectivity and can be measured via tracer studies in animals (Markov et al., 2012) \nocite{markov2012weighted}
and diffusion based magnetic resonance imaging (MRI) technologies in humans (Johansen-Berg \& Rushworth, 2009). \nocite{johansen2009using} Functional connectivity refers to the temporal statistical relationship between spatially distinct brain regions and is usually inferred from the time series coherence/correlation in brain activity between regions of interest (Park \& Friston, 2013). \nocite{park2013structural} In humans, both structural connectivity and functional connectivity are commonly represented as undirected graphs because directionality information is usually not available via the measurement process.
In contrast, effective connectivity is represented as a directed graph where the edge direction in the graph represents the influence of one node on another, inferred from a particular model of causal dynamics (Friston et al., 2003). \nocite{friston2003dynamic} This paper focuses on human functional connectivity measured via fMRI.

\subsection{Observations}
Resting state functional connectivity data is often represented as a graph, where a graph is a collection of nodes and edges. The
nodes represent meaningful units of information processing, such as brain areas, and the edges, also called functional links, represent the interdependency between the nodes. Representation of functional brain organization data as a graph allows us to borrow constructs from the field of network science (Newman, 2010) \nocite{newman2010networks} and has inspired new ways of thinking about brain organization. For example, functional networks have been shown to display features such as short characteristic path length (Achard et al., 2006), \nocite{achard2006resilient} local clustering (Achard et al., 2006), \nocite{achard2006resilient} modular structure (Ferrarini et al., 2009), \nocite{ferrarini2009hierarchical} exponentially truncated degree distributions (Eguiluz et al., 2005; Achard et al., 2006), \nocite{eguiluz2005scale} \nocite{achard2006resilient} overabundance of short distance connections (Salvador et al., 2005) \nocite{salvador2005neurophysiological} and the presence of different types of hubs (Buckner et al., 2009; Tomasi \& Volkow, 2011). \nocite{buckner2009cortical}\nocite{tomasi2011functional} Additionally, it has also been observed that the presence and strength of structural connectivity shapes functional connectivity, (Honey et al., 2009) \nocite{honey2009predicting} although it is also possible to observe functional connectivity without the presence of direct structural connections (Vincent et al., 2007). \nocite{vincent2007intrinsic} Lastly, as mentioned earlier, the main value of resting state functional connectivity lies in regions of the brain which share function displaying correlation in the low frequency components of the BOLD signal. Functional neuroanatomical systems which display this type of correlation are often called resting state networks, and numerous such networks have been found such as those related to memory (Vincent et al., 2006), \nocite{vincent2006coherent} visual (Cordes et al, 2000), \nocite{cordes2000mapping} language (Hampson et al., 2002), \nocite{hampson2002detection} auditory (Cordes et al., 2000), \nocite{cordes2000mapping} salience (Seeley et al., 2007), \nocite{seeley2007dissociable} motor (Biswal et al., 1995) \nocite{biswal1995functional} and control (Vincent et al., 2008) \nocite{vincent2008evidence} systems.

\subsection{Computational Principles: Moving from Knowledge to Understanding}
As seen above, a core set of observations for functional brain networks has already been collected. As we continue to collect more knowledge about these networks, it is also important to step back and ask what type of rules or computational principles (Churchland \& Sejnowski, 1992) \nocite{churchland1992computational} might give rise to these observations. Very little is known about these rules. Vertes et al. (2012)\nocite{vertes2012simple}
is a recent and rare attempt at formalizing this question. The approach taken by Vertes et al. (2012) assumed a parameterized probability model that the probability of a functional link between two regions of interest in a brain network is present or absent. It was assumed that the likelihood of an edge is functionally dependent upon the model's parameters and the manner in which that edge is connected to other edges in the network through specific topological features of the entire network. An ``energy function'' was then constructed which had the property that a global minimizer of the energy function
corresponded to a solution identifying which topological features are predictive of the likelihood of an edge, as well as how those topological
properties are predictive. The basic concept of the energy function was to encapsulate the idea that the difference in topological indices (clustering, efficiency, modularity and degree distribution) between generated model networks and observed brain networks should be
minimized. Thus, the parameters of their model could be estimated by minimizing an energy function. The goal of this analysis was to not only make predictions about the likelihood of coupling, but more importantly to identify which topological features are relevant to functional brain networks. After the parameters were
estimated, networks were generated using the probability model with the learned parameter values. Validation consisted of checking if the topological indices calculated on generated model networks matched those of the observed brain networks. Their main observation was that a probability law for a functional link between two regions, where there is competition between a distance penalty term (Euclidean distance used as a proxy for wiring cost) and a topological feature which favors links between regions sharing input nodes,  can capture many known topological indices of these networks.  It is important to note, as also emphasized in Vertes et al. (2012), that such a rule (despite the biological plausibility) is computationally sufficient but not necessary to explain the observations. Multiple semantically similar energy functions were constructed, and the main results did not change as a function
of the form of the energy function. Lastly, they showed that the difference in networks of healthy controls versus a schizophrenic group is reflected
in differing parameter values. 

As expected from any early attempt at a hard problem, the approach of Vertes et al. (2012) leaves room for improvement. From a neuroscience
perspective, it is important to realize that any generated model network which displays topological indices similar to observed brain networks,
is not necessarily sufficiently grounded in the known neuroscience. This is  because multiple different configurations of a network can realize the same index.
For example, let's assume that there is an index that quantifies the modular organization of a network. If the value of this index for generated model networks matches that of observed brain networks, further validation would still be  required to ascertain that the modular organization of the generated network does in fact follow the known modules of the human brain such as the visual system, sensorimotor system, auditory system etc. There are also possibilities for methodological improvements. Due to the lack of a principled choice of an energy function, multiple different semantically similar energy functions were tested to increase confidence in the results, and it was shown that the main results do not change qualitatively as a function of the energy function. Despite this effort, it is quite conceivable that energy functions which were not tested, but are semantically similar, might qualitatively change the main results. There is another very crucial methodological limitation. The energy function implicitly informed the model of the topological indices which the generative model should learn, since the goal of the energy function was to minimize the difference in value of topological indices between generated and observed brain networks. Since these same topological indices were then used for the validation of the generated networks, there is an inherent circularity.  This was explicitly recognized in Vertes et al. (2012) and hence during the validation phase the generated model networks were also compared to a set of observed brain networks which had not been used in the training phase. Despite this effort, the confidence in the results would have been greater if there was a way to circumvent this circularity. Lastly, the difference in parameter values between the healthy and schizophrenic group were shown to be qualitatively different, but it is not certain if this difference is real or due to sampling error since no error bounds were provided.

\section{Goal}
In fMRI, blood flow variables are a proxy for neuronal activity variables. Additionally, it is well known that thermal noise, system noise, and physiological artifacts can all affect the fMRI signal (Huettel et al., 2004). Hence, resting state functional connectivity fMRI data can be considered a noisy representation of the ground truth functional network structure. Thus, the first goal of this paper is to develop a methodology which can learn a better representation of the initial fMRI data. Additionally, a large number of robust observations has been collected for resting state networks. This is truly remarkable considering the large variance in scanner types, imaging parameters and pre-processing pipelines. Yet, apart from Vertes et al. (2012), we do not know of other formal attempts at moving from these observations to formal explanations which can account for these observations.  Any new field of inquiry requires a minimal number of observations before it is worthwhile to dive into theoretical endeavors. This paper argues that the threshold of minimal number of observations has been reached, and takes the philosophical viewpoint that unless we move from observations to theory, we will not be able to move from knowledge to understanding. But, this is easier said than done. The limitations of the pioneering efforts of Vertes et al. (2012), point toward a lack of a principled approach to this problem. Thus, the second goal of this paper is to develop a methodology which can address these limitations.  To concurrently achieve the above stated goals, this paper develops a statistical machine learning framework for generative modeling of functional connectivity networks. We specifically call this a \emph{framework} because we are not developing a single model. Rather, we are providing a scaffolding which can be used to postulate and test multiple models which represent multiple competing theories of functional connectivity. This is achieved by using ``feature functions" (explained in Section 4), which allows the researcher to postulate different theories of functional connectivity while minimally changing the mathematics.

The framework is based on a Hidden Markov Random Field (HMRF). A HMRF is a latent variable probabilistic graphical model that can represent a joint probability distribution over random variables, some of which are observable and some of which are latent. As eluded above, in fMRI, one can conceptualize two variables of interest because hemodynamic response is a proxy for neuronal activity. More specifically, the first set of variables can represent the non-observable (latent/hidden) neuronal activity, and the second set of variables can represent the hemodynamic response. The distinction between observed variables and latent variables in a HMRF can be used advantageously to explicitly recognize that much theoretical work remains to be carried out to understand the coupling relationship between neuronal activity and hemodynamic response in fMRI (Logothetis, 2008), \nocite{logothetis2008we} and hence prevent the confounding of observed correlation values (inferred from the hemodynamic response) and hidden functional connectivity network structure. The word ``Markov'' refers to the idea that each random variable is not independent, but rather is dependent upon a subset of the other random variables, thereby facilitating the modeling of conditional dependence structures between the random variables. The probability rule for an edge in the hidden network structure can then be postulated based on a researchers theory of what influences functional connectivity. Once the model is explicitly represented, the parameters of the model can be learned in a maximum likelihood estimation (MLE) framework. In MLE, the goal of learning is to find the parameters which maximize the likelihood of the observed dataset. Thus, the objective function (energy function) in a MLE framework does not implicitly inform the model of the desired topological indices. Once the parameter values have been learned, functional brain networks can be generated by sampling from the joint probability distribution. The generated networks can be validated by checking if they are able to capture the known organizational features of the brain without limiting them a priori to a few topological indices. The tools of model selection criteria can be used to compare multiple models, model misspecification tests can be used to check if a postulated model is capable of representing the data generating process, and hypothesis testing can be carried out to test if parameter values change across experimental conditions (such as healthy vs.schizophrenic). Lastly, this estimation framework also allows us to state guarantees on parameter estimates such as its consistency, efficiency, asymptotic distribution and error bounds (Greene, 2003). \nocite{greene2003econometric} 

It is necessary to highlight that the HMRF framework brings with it two important innovations. First, the importance of the ability to make the distinction between observed blood flow variables and hidden neuronal variables cannot be overstated. Currently, despite the general acceptance that blood flow variables are a very high level approximation of neuronal activity variables, it is often assumed by the practitioner that there are no available methods which can distinguish between these two variables. Secondly, the importance of the simple ``Markov" idea also cannot be overstated. For example, thresholding is a common approach to infer the true network structure given the observed fMRI correlational data. In one type of thresholding, the top ``x" percent of the strongest functional links (where x is usually a number between 2 and 10) are considered important and the rest of the links are pruned. This is a ``poor man's approach" to representation learning. The implicit assumption here is that weak links are not important, and that functional links are conditionally independent. Such assumptions are not biologically accurate (see Bassett et al., (2011) for an explicit example where weak links have been shown to be important), and yet thresholding remains a dominant method to infer network structure due to the absence of other methods which can make better assumptions. The HMRF framework tackles both the above problems.

\section{Hidden Markov Random Field Framework: Mathematical Development}
After data collection and pre-processing, the final form of the data is an adjacency matrix to which the Hidden Markov Random Field (HMRF) framework will be applied. It is assumed that the brain has already been parcellated into appropriate regions of interest (nodes), and that the interdependency between the nodes (i.e., edges representing functional connectivity) has been quantified as Pearson's correlation coefficient. Pearson's correlation coefficient is the most commonly used measure of functional connectivity. Additionally, instead of postulating specific computational principles, we introduce the framework more generally via ``feature functions" so that researchers can postulate and test their own theories of functional connectivity. 

The thinking behind using feature functions goes as follows. It is not sufficient to just identify an unresolved problem, build a tool to tackle the problem which clearly presents all the modeling assumptions, and is overall sound in its mathematical formulation. One has to aspire to go beyond, and introduce the tool to the practitioner in a manner where ideas can be quickly tested. The feature function aids this very important, and often neglected goal, i.e., presenting a new mathematical tool to the practitioner where ease of use is fundamentally valued. The basic idea of a ``feature function" is simple. The feature function is a collection of factors that the researcher postulates as being the factors which influence functional connectivity between any two nodes. For example, one could postulate the same factor that Vertes et al. (2012) showed to be important, i.e., a feature which counts the number of nearest neighbors (i.e. nodes) in common between the the two nodes that an edge connects. Another type of feature could be a ``system membership" feature which represents if an edge connecting two nodes is within the same system or between systems. A practitioner is more interested in testing various hypotheses about factors/features which affect functional connectivity, as compared to the mathematical formulation of the tool. Hence, all they need to do to is simply interact with the feature function module in the software package, and test multiple models quickly, while the mathematical machinery consisting of undirected graphs, latent variables, learning algorithm, sampling algorithms etc., does not compete for their focus.

\subsection{Data Generating Process}
The statistical environment of the learning machine is conceptualized
as consisting of two variables of interest. The first set of variables, $\mathbf{x}\in\{0,1\}^{d}$,
represents the hidden functional network structure. The $i^{th}$individual element of $\mathbf{x}$, $x_{i}$, takes on the value of 1 if the $i^{th}$ edge in the hidden functional network is present and takes on the value of 0 otherwise. The second set of variables, $\mathbf{y\in} {\cal R}^d$, represents the observed Fisher transformed correlation in BOLD signal between brain regions. The $i^{th}$individual element of $\mathbf{y}$, $y_{i}$, is the observed Fisher transformed correlation in BOLD signal between the two brain regions that the $i^{th}$ edge connects.
The training data consists of $n$ data records (1 data point per subject). The data set ${\cal D}_n$ is defined by: ${\cal D}_n \equiv \{ {\bf y}_1, \ldots, {\bf y}_n \}$, since ${\bf x}_1, \ldots, {\bf x}_n$ are never directly observable. Formally, the data set ${\cal D}_n$ is a realization of a random sample which is a stochastic sequence of $n$ independent and identically distributed $d$-dimensional random vectors with common DGP $P^{0}$.

\subsection{Notation}

\begin{itemize}
\item $d$ : total number of edges in network
\item $n$ : total number of networks/subjects
\item $\mathbf{x}$ : vector representing the hidden variables in the network,
$\mathbf{x}\in\{0,1\}^{d}$
\item $x_{i}$ : $i^{th}$individual element of $\mathbf{x}$, $i=1\ldots d$
\item $\mathcal{N}_{i}(\mathbf{x})$: neighborhood graph of $x_{i}$

\item $\mathbf{y}$ : vector representing the observable variables, $\mathbf{y\in} {\cal R}^d$
\item $y_{i}$ : $i^{th}$individual element of $\mathbf{y}$, $i=1\ldots d$
\item $t$ : index for subject, $t=1\ldots n$
\item ${\bf f}_{i}$ : a function ${\bf f}_{i} :  {\cal R}^d \rightarrow {\cal R}^{k+1}$
which is defined such that ${\bf f}_{i}({\bf x})$ is a nonlinear hidden
representation of ${\bf x}$.

\item ${D}_{i}$ : a function ${D}_{i} :  {\cal R}^d \rightarrow {\cal R}$
which is defined such that ${D}_{i}({\bf x})$ is the Euclidean distance between the two nodes that ${x}_{i}$ connects. Euclidean distance is used as a proxy for wiring cost.

\end{itemize}

\subsection{Probability Model}
The probabilistic modeling assumptions are as follows.

It is assumed that the probability density of a Fisher transformed correlation between BOLD activity in two brain regions, , $y_{i}$, 
is Gaussian which is conditionally dependent upon the presence/absence
of a latent functional link, i.e., $x_{i}$. This is justified by the theoretical result that the sampling distribution of a Fisher transformed correlation coefficient is Gaussian (Fisher, 1915; Fisher, 1921).\nocite{fisher1921probable} \nocite{fisher1915frequency} More specifically, it is assumed that $y_{i}$ is distributed as a Gaussian with mean $\alpha_{1}$
and variance $\sigma_{1}$ when $x_{i}=1,$ and as a Gaussian with
mean $\alpha_{0}$ and variance $\sigma_{0}$ when $x_{i}=0.$ $\alpha_{1}$,
$\alpha_{0},$ $\sigma_{1}$ and $\sigma_{0}$ are treated as free
parameters. This can be compactly represented as:

\begin{equation}
p(y_{i}\mid x_{i})=(\sqrt{2\pi}(\sigma_{1}x_{i}+\sigma_{0}(1-x_{i}))^{-1}\exp(-\frac{(y_{i}-(\alpha_{1}x_{i}+\alpha_{0}(1-x_{i}))^{2}}{2(\sigma_{1}x_{i}+\sigma_{0}(1-x_{i}))^{2}}). 
\end{equation}

It is assumed that the $y_{i}'s$ are conditionally independent given
the $x_{i}'s$. $\mathbf{m}$ is a $d$-dimensional
vector with $i^{th}$ element $(x_{i}\alpha_{1}+(1-x_{i})\alpha_{0})$.
$\mathbf{C}$ is a $d$ x $d$ dimensional covariance matrix with $ii^{th}$
diagonal element $(x_{i}\sigma_{1}+(1-x_{i})\sigma_{0})$ and all off
diagonal elements $0's$. In addition, 

\begin{equation}
p({\bf y} | {\bf x}; {\bm \theta}) = \prod_i p(y_i | x_i; {\bm \theta})=((\sqrt{2\pi})^{d}det(\mathbf{C}))^{-1}exp(-(1/2)(\mathbf{y}\mathbf{-}\mathbf{m})^{T}\mathbf{C}^{-1}(\mathbf{y}\mathbf{-}\mathbf{m}))
\end{equation}
where
\begin{displaymath}
\bm{\theta}=[\alpha_{0} \;\; \alpha_{1} \;\;  \sigma_{0} \;\;  \sigma_{1}]^{T}
\end{displaymath}

which is the Multivariate Gaussian Distribution with mean vector $\mathbf{m}$
and covariance matrix $\mathbf{C}$.

A slightly different, but semantically more meaningful approach is
taken for the postulation of $p(\mathbf{x}).$ Instead of directly
postulating $p(\mathbf{x}),$ a potential function is first postulated
for the field $\mathbf{x}$. The potential function is a natural
way to formally capture a researcher's intuition about the hidden
network structure. The basic idea is that configurations of $\mathbf{x}$
which have higher probability are given lower potential. Another key
idea is that of a neighborhood graph. The neighborhood of an edge
$x_{i}$ is represented as $\mathcal{N}_{i}(\mathbf{x})$ and consists
of a collection of other edges on which $x_{i}$ is conditionally
dependent.  For example, one could assume that the neighborhood of an edge consists
of all the other edges which share nodes with that edge. Once a potential
function for the field $\mathbf{x}$ has been postulated, the Hammersley-Clifford
Theorem (Golden, 1996) \nocite{golden1996mathematical} can be invoked to write down a legitimate probability
density over $\mathbf{x}$, i.e., $p(\mbox{\textbf{x}) .}$ A step
by step procedure for deriving $p(\mathbf{x})$is now presented.
\begin{itemize}
\item First a potential function for a single edge, $V(x_{i}|\mathcal{N}_{i}(\mathbf{x}))$, is postulated. The quantity ``V" may be interpreted as the amount
of ``support" or ``evidence" that an edge is present, with smaller values of $V$ corresponding to increased support or evidence. The value of $V$ depends upon a
set of features or topological properties that is represented in the
 {\em feature function} $\mathbf{f}_{i}$. The feature function ${\bf f}_i : {\cal R}^d \rightarrow {\cal R}^{k+1}$
is defined such that ${\bf f}_i({\bf x})$ is a collection of $k+1$ features/factors which are properties of the graph specified by
${\bf x}$, and postulated by the researcher to influence functional connectivity.  For example, as used with good results in Vertes et al. (2012), a topological feature which counts the number of nearest neighbors (i.e. nodes) in common between the the two nodes that an edge connects, could be one of the elements of the feature function. Similarly, $k$ other features can be hypothesized. Additionally, regularization can also be introduced. For example, one could penalize the likelihood of an edge as a function of wiring cost, $D_{i}$, where
 ${D}_{i} :  {\cal R}^d \rightarrow {\cal R}$ is defined such that ${D}_{i}({\bf x})$ is the Euclidean distance between the two nodes that ${x}_{i}$ connects. Euclidean distance as a proxy for wiring cost has previously been shown to yield good results (Vertes et al., 2012).

\begin{equation}
V(x_{i} | \mathcal{N}_{i}(\mathbf{x}) ; \bm{\beta})=-\bm{\beta}^{T}\mathbf{f}_{i}+||\bm{\beta}||^{2}D_{i}
\end{equation}

where
\begin{displaymath}
\bm{\beta}=[\begin{array}{ccccc}
\beta_{0} & \beta_{1} & \beta_{2} & \ldots & \beta_{k}]\end{array}^{T}
\end{displaymath}

(The functional form of the potential function can be justified using the ``economy" theory of brain organization which states that brain organization negotiates a trade-off between increasing adaptive value and minimizing wiring costs. See Bullmore \& Sporns (2012) for extensive neuroscience evidence.) \nocite{bullmore2012economy}

\end{itemize}

\ 
\begin{itemize}
\item The potential function for the entire field can then be written as
a summation of the individual potential functions.

\begin{equation}
V(\mathbf{x}; \bm{\beta})=-\bm{\beta}^{T}{\displaystyle \sum_{i}\mathbf{f}_{i}+}||\bm{\beta}||^{2}{\displaystyle \sum_{i}D_{i}}
\end{equation}
The quantity $V({\bf x}; {\bm \beta})$ is a monotonically decreasing function of the evidence supporting the frequency of occurrence of  ${\bf x}$.
\end{itemize}

\ 
\begin{itemize}
\item The joint density function over the field $\mathbf{x}$ can then be
written as:

\begin{equation}
p(\mathbf{x}|\bm{\beta})=\frac{1}{Z_{\mathbf{x}}}exp(-V(\mathbf{x}))=\frac{1}{Z_{\mathbf{x}}}exp(\bm{\beta}^{T}{\displaystyle \sum_{i}\mathbf{f}_{i}-}||\bm{\beta}||^{2}{\displaystyle \sum_{i}D_{i})}
\end{equation}
where

\begin{displaymath}
Z_{\mathbf{x}}={\displaystyle \sum_{\mathbf{x}\in\{0,1\}^{d}}exp(\bm{\beta}^{T}{\displaystyle \sum_{i}\mathbf{f}_{i}-}||\bm{\beta}||^{2}{\displaystyle \sum_{i}D_{i})}}
\end{displaymath}

\end{itemize}
\ 

Using the definition of conditional probability, the joint density
over $\mathbf{x}$ and $\mathbf{y}$ can now be written as:

\begin{equation}
p({\bf x}, {\bf y} | {\bm \beta}, {\bm \theta}) = p({\bf y} | {\bf x}; {\bm \theta}) P({\bf x} | {\bm \beta})
\end{equation}

\subsection{Learning and Inference}
The parameter vectors ${\bm \theta}$ and ${\bm \beta}$ are
simultaneously  estimated using the method of maximum likelihood estimation
by minimizing the objective function:
\begin{equation}
\label{likelihoodfunk}
\ell_n({\bm \theta}, {\bm \beta}) =
-(1/n) \sum_{t=1}^n \log \sum_q p( {\bf y}^t |  {\bf x}^q ;{\bm \theta}) p({\bf x}^q | {\bm \beta})
\end{equation}
which maximizes the likelihood of the observed data ${\bf y}_1, \ldots, {\bf y}_n$.
The evaluation of the summation in this objective function is computationally intractable. However,
contrastive-divergence type of methods (Bengio \& Delalleau, 2009) \nocite{bengio2009justifying} in conjunction with the Metropolis-Hastings method (Hastings, 1970) \nocite{hastings1970monte} for sampling from the probability mass function $p({\bf x} | {\bm \beta})$, can be used to make the problem computationally tractable.
 
A Metropolis-Hasting method (Bishop, 2006) \nocite{Bishop:2006:PRM:1162264} can be used to generate latent variable values given the observed values, for the purpose of inferring the latent functional network structure from the noisy fMRI data. In addition, model selection criteria such as Bayes Information Criterion (BIC) (Bishop, 2006) \nocite{Bishop:2006:PRM:1162264} can then be used as the basis for comparing competing hypotheses. 

\section{Discussion}
This paper starts with two basic premises. 1) Resting state functional connectivity fMRI data is a noisy representation of the functional organization of the brain. 2) Researchers have collected a robust set of observations about functional connectivity networks and yet very little is known about the type of computational principles which operate in these networks and give rise to the observations. Based on these two premises, a Hidden Markov Random Field framework for generative modeling of resting state functional connectivity networks has been proposed. This framework serves the joint purpose of learning representations of the observable data, while testing computational principles which operate in these networks.  

In the proposed framework, two types of functional connectivity variables are conceptualized. One set of variables represents the observable functional connectivity blood flow fMRI data, and the second set of variables represents the hidden functional connectivity network structure. It is assumed that an observed value is conditionally dependent upon the presence or absence of a latent functional connectivity link. More specifically, it is assumed that an observable value is a mixture of two Gaussian distributions, one signifying the presence of a latent functional link and the other the absence. The probability of a latent functional link itself depends upon properties in the hidden functional network structure which are hypothesized by a researcher to be predictive of the presence or absence of a functional link.

The theoretical contributions of this paper can be summarized as follows.

\begin{enumerate}
  \item A probabilistic framework for resting state functional connectivity fMRI analysis which makes an explicit distinction between observable variables (fMRI blood flow variables) and latent/hidden variables (neuronal activity variables).
  \item  Provides a method which can model the conditional dependence structure between functional links, thereby assuming a more biologically plausible architecture than the commonly made conditional independence assumption.
  \item Provides a method to learn a representation of the latent neuronal functional network structure from noisy fMRI data.
  \item Provides a method for testing multiple competing hypotheses of functional connectivity principles.
  \item The approach involves the use of  ``feature mapping functions" which corresponds to different hypotheses regarding what properties of the latent functional connectivity network are predictive of the presence or absence of a functional link.
  \item The number of free parameters is relatively small. There are only $k+1$ free parameters for specifying the latent structure and only 4 parameters specifying the observable structure.
  \item All parameters of the model have semantically interpretable properties.
  \item The probabilistic framework provides guidance regarding what objective function should be used to ``match" estimated free parameters to the observed data. Specifically, maximum likelihood methods are used rather than ad hoc methods of previous approaches.
  \item The probabilistic framework provides a methodology for estimating the likelihood of a model given the observed data. This can be used as the basis for a model selection criteria for comparing competing hypotheses. For example, deciding which of two probability models is most probable given the observed data.
  \
\end{enumerate}

\bibliography{iclr2015}

\begin{thebibliography}{47}
\providecommand{\natexlab}[1]{#1}
\providecommand{\url}[1]{\texttt{#1}}
\expandafter\ifx\csname urlstyle\endcsname\relax
  \providecommand{\doi}[1]{doi: #1}\else
  \providecommand{\doi}{doi: \begingroup \urlstyle{rm}\Url}\fi

\bibitem[Achard et~al.(2006)Achard, Salvador, Whitcher, Suckling, and
  Bullmore]{achard2006resilient}
Achard, Sophie, Salvador, Raymond, Whitcher, Brandon, Suckling, John, and
  Bullmore, Ed.
\newblock A resilient, low-frequency, small-world human brain functional
  network with highly connected association cortical hubs.
\newblock \emph{The Journal of Neuroscience}, 26\penalty0 (1):\penalty0 63--72,
  2006.

\bibitem[Allen et~al.(2012)Allen, Damaraju, Plis, Erhardt, Eichele, and
  Calhoun]{allen2012tracking}
Allen, Elena~A, Damaraju, Eswar, Plis, Sergey~M, Erhardt, Erik~B, Eichele, Tom,
  and Calhoun, Vince~D.
\newblock Tracking whole-brain connectivity dynamics in the resting state.
\newblock \emph{Cerebral cortex}, pp.\  bhs352, 2012.

\bibitem[Beckmann \& Smith(2005)Beckmann and Smith]{beckmann2005tensorial}
Beckmann, Christian~F and Smith, Stephen~M.
\newblock Tensorial extensions of independent component analysis for
  multisubject fmri analysis.
\newblock \emph{Neuroimage}, 25\penalty0 (1):\penalty0 294--311, 2005.

\bibitem[Bengio \& Delalleau(2009)Bengio and Delalleau]{bengio2009justifying}
Bengio, Yoshua and Delalleau, Olivier.
\newblock Justifying and generalizing contrastive divergence.
\newblock \emph{Neural Computation}, 21\penalty0 (6):\penalty0 1601--1621,
  2009.

\bibitem[Bishop(2006)]{Bishop:2006:PRM:1162264}
Bishop, Christopher~M.
\newblock \emph{Pattern Recognition and Machine Learning (Information Science
  and Statistics)}.
\newblock Springer-Verlag New York, Inc., Secaucus, NJ, USA, 2006.
\newblock ISBN 0387310738.

\bibitem[Biswal et~al.(1995)Biswal, Zerrin~Yetkin, Haughton, and
  Hyde]{biswal1995functional}
Biswal, Bharat, Zerrin~Yetkin, F, Haughton, Victor~M, and Hyde, James~S.
\newblock Functional connectivity in the motor cortex of resting human brain
  using echo-planar mri.
\newblock \emph{Magnetic resonance in medicine}, 34\penalty0 (4):\penalty0
  537--541, 1995.

\bibitem[Buckner et~al.(2009)Buckner, Sepulcre, Talukdar, Krienen, Liu, Hedden,
  Andrews-Hanna, Sperling, and Johnson]{buckner2009cortical}
Buckner, Randy~L, Sepulcre, Jorge, Talukdar, Tanveer, Krienen, Fenna~M, Liu,
  Hesheng, Hedden, Trey, Andrews-Hanna, Jessica~R, Sperling, Reisa~A, and
  Johnson, Keith~A.
\newblock Cortical hubs revealed by intrinsic functional connectivity: mapping,
  assessment of stability, and relation to alzheimer's disease.
\newblock \emph{The Journal of Neuroscience}, 29\penalty0 (6):\penalty0
  1860--1873, 2009.

\bibitem[Bullmore \& Sporns(2012)Bullmore and Sporns]{bullmore2012economy}
Bullmore, Ed and Sporns, Olaf.
\newblock The economy of brain network organization.
\newblock \emph{Nature Reviews Neuroscience}, 13\penalty0 (5):\penalty0
  336--349, 2012.

\bibitem[Churchland \& Sejnowski(1992)Churchland and
  Sejnowski]{churchland1992computational}
Churchland, Patricia~Smith and Sejnowski, Terrence~J.
\newblock \emph{The computational brain.}
\newblock The MIT press, 1992.

\bibitem[Cordes et~al.(2000)Cordes, Haughton, Arfanakis, Wendt, Turski, Moritz,
  Quigley, and Meyerand]{cordes2000mapping}
Cordes, Dietmar, Haughton, Victor~M, Arfanakis, Konstantinos, Wendt, Gary~J,
  Turski, Patrick~A, Moritz, Chad~H, Quigley, Michelle~A, and Meyerand,
  M~Elizabeth.
\newblock Mapping functionally related regions of brain with functional
  connectivity mr imaging.
\newblock \emph{American Journal of Neuroradiology}, 21\penalty0 (9):\penalty0
  1636--1644, 2000.

\bibitem[Cordes et~al.(2001)Cordes, Haughton, Arfanakis, Carew, Turski, Moritz,
  Quigley, and Meyerand]{cordes2001frequencies}
Cordes, Dietmar, Haughton, Victor~M, Arfanakis, Konstantinos, Carew, John~D,
  Turski, Patrick~A, Moritz, Chad~H, Quigley, Michelle~A, and Meyerand,
  M~Elizabeth.
\newblock Frequencies contributing to functional connectivity in the cerebral
  cortex in resting-state data.
\newblock \emph{American Journal of Neuroradiology}, 22\penalty0 (7):\penalty0
  1326--1333, 2001.

\bibitem[Cowen et~al.(2014)Cowen, Chun, and Kuhl]{cowen2014neural}
Cowen, Alan~S, Chun, Marvin~M, and Kuhl, Brice~A.
\newblock Neural portraits of perception: reconstructing face images from
  evoked brain activity.
\newblock \emph{Neuroimage}, 94:\penalty0 12--22, 2014.

\bibitem[De~Martino et~al.(2008)De~Martino, Valente, Staeren, Ashburner,
  Goebel, and Formisano]{de2008combining}
De~Martino, Federico, Valente, Giancarlo, Staeren, No{\"e}l, Ashburner, John,
  Goebel, Rainer, and Formisano, Elia.
\newblock Combining multivariate voxel selection and support vector machines
  for mapping and classification of fmri spatial patterns.
\newblock \emph{Neuroimage}, 43\penalty0 (1):\penalty0 44--58, 2008.

\bibitem[Eguiluz et~al.(2005)Eguiluz, Chialvo, Cecchi, Baliki, and
  Apkarian]{eguiluz2005scale}
Eguiluz, Victor~M, Chialvo, Dante~R, Cecchi, Guillermo~A, Baliki, Marwan, and
  Apkarian, A~Vania.
\newblock Scale-free brain functional networks.
\newblock \emph{Physical review letters}, 94\penalty0 (1):\penalty0 018102,
  2005.

\bibitem[Esteban et~al.(2014)Esteban, Wollny, Gorthi, Ledesma-Carbayo, Thiran,
  Santos, and Bach-Cuadra]{esteban2014mbis}
Esteban, Oscar, Wollny, Gert, Gorthi, Subrahmanyam, Ledesma-Carbayo,
  Mar{\'\i}a-J, Thiran, Jean-Philippe, Santos, Andr{\'e}s, and Bach-Cuadra,
  Meritxell.
\newblock Mbis: Multivariate bayesian image segmentation tool.
\newblock \emph{Computer methods and programs in biomedicine}, 115\penalty0
  (2):\penalty0 76--94, 2014.

\bibitem[Ferrarini et~al.(2009)Ferrarini, Veer, Baerends, van Tol, Renken,
  van~der Wee, Veltman, Aleman, Zitman, Penninx,
  et~al.]{ferrarini2009hierarchical}
Ferrarini, Luca, Veer, Ilya~M, Baerends, Evelinda, van Tol, Marie-Jos{\'e},
  Renken, Remco~J, van~der Wee, Nic~JA, Veltman, Dirk, Aleman, Andr{\'e},
  Zitman, Frans~G, Penninx, Brenda~WJH, et~al.
\newblock Hierarchical functional modularity in the resting-state human brain.
\newblock \emph{Human brain mapping}, 30\penalty0 (7):\penalty0 2220--2231,
  2009.

\bibitem[Fisher(1915)]{fisher1915frequency}
Fisher, Ronald~A.
\newblock Frequency distribution of the values of the correlation coefficient
  in samples from an indefinitely large population.
\newblock \emph{Biometrika}, pp.\  507--521, 1915.

\bibitem[Fisher et~al.(1921)]{fisher1921probable}
Fisher, Ronald~Aylmer et~al.
\newblock On the" probable error" of a coefficient of correlation deduced from
  a small sample.
\newblock \emph{Metron}, 1:\penalty0 3--32, 1921.

\bibitem[Friston et~al.(2003)Friston, Harrison, and Penny]{friston2003dynamic}
Friston, Karl~J, Harrison, Lee, and Penny, Will.
\newblock Dynamic causal modelling.
\newblock \emph{Neuroimage}, 19\penalty0 (4):\penalty0 1273--1302, 2003.

\bibitem[Golden(1996)]{golden1996mathematical}
Golden, Richard~M.
\newblock \emph{Mathematical methods for neural network analysis and design}.
\newblock MIT Press, 1996.

\bibitem[Greene(2003)]{greene2003econometric}
Greene, William~H.
\newblock \emph{Econometric analysis}.
\newblock Pearson Education India, 2003.

\bibitem[Hamilton et~al.(2011)Hamilton, Chen, Thomason, Schwartz, and
  Gotlib]{hamilton2011investigating}
Hamilton, J~Paul, Chen, Gang, Thomason, Moriah~E, Schwartz, Mirra~E, and
  Gotlib, Ian~H.
\newblock Investigating neural primacy in major depressive disorder:
  multivariate granger causality analysis of resting-state fmri time-series
  data.
\newblock \emph{Molecular psychiatry}, 16\penalty0 (7):\penalty0 763--772,
  2011.

\bibitem[Hampson et~al.(2002)Hampson, Peterson, Skudlarski, Gatenby, and
  Gore]{hampson2002detection}
Hampson, Michelle, Peterson, Bradley~S, Skudlarski, Pawel, Gatenby, James~C,
  and Gore, John~C.
\newblock Detection of functional connectivity using temporal correlations in
  mr images.
\newblock \emph{Human brain mapping}, 15\penalty0 (4):\penalty0 247--262, 2002.

\bibitem[Hastings(1970)]{hastings1970monte}
Hastings, W~Keith.
\newblock Monte carlo sampling methods using markov chains and their
  applications.
\newblock \emph{Biometrika}, 57\penalty0 (1):\penalty0 97--109, 1970.

\bibitem[Honey et~al.(2009)Honey, Sporns, Cammoun, Gigandet, Thiran, Meuli, and
  Hagmann]{honey2009predicting}
Honey, CJ, Sporns, O, Cammoun, Leila, Gigandet, Xavier, Thiran, Jean-Philippe,
  Meuli, Reto, and Hagmann, Patric.
\newblock Predicting human resting-state functional connectivity from
  structural connectivity.
\newblock \emph{Proceedings of the National Academy of Sciences}, 106\penalty0
  (6):\penalty0 2035--2040, 2009.

\bibitem[Honnorat et~al.(2015)Honnorat, Eavani, Satterthwaite, Gur, Gur, and
  Davatzikos]{honnorat2015grasp}
Honnorat, N, Eavani, H, Satterthwaite, TD, Gur, RE, Gur, RC, and Davatzikos, C.
\newblock Grasp: Geodesic graph-based segmentation with shape priors for the
  functional parcellation of the cortex.
\newblock \emph{NeuroImage}, 106:\penalty0 207--221, 2015.

\bibitem[Johansen-Berg \& Rushworth(2009)Johansen-Berg and
  Rushworth]{johansen2009using}
Johansen-Berg, Heidi and Rushworth, Matthew~FS.
\newblock Using diffusion imaging to study human connectional anatomy.
\newblock \emph{Annual review of neuroscience}, 32:\penalty0 75--94, 2009.

\bibitem[Kim et~al.(2007)Kim, Zhu, Chang, Bentler, and Ernst]{kim2007unified}
Kim, Jieun, Zhu, Wei, Chang, Linda, Bentler, Peter~M, and Ernst, Thomas.
\newblock Unified structural equation modeling approach for the analysis of
  multisubject, multivariate functional mri data.
\newblock \emph{Human Brain Mapping}, 28\penalty0 (2):\penalty0 85--93, 2007.

\bibitem[LaConte et~al.(2005)LaConte, Strother, Cherkassky, Anderson, and
  Hu]{laconte2005support}
LaConte, Stephen, Strother, Stephen, Cherkassky, Vladimir, Anderson, Jon, and
  Hu, Xiaoping.
\newblock Support vector machines for temporal classification of block design
  fmri data.
\newblock \emph{NeuroImage}, 26\penalty0 (2):\penalty0 317--329, 2005.

\bibitem[Logothetis(2008)]{logothetis2008we}
Logothetis, Nikos~K.
\newblock What we can do and what we cannot do with fmri.
\newblock \emph{Nature}, 453\penalty0 (7197):\penalty0 869--878, 2008.

\bibitem[Markov et~al.(2012)Markov, Ercsey-Ravasz, Gomes, Lamy, Magrou, Vezoli,
  Misery, Falchier, Quilodran, Gariel, et~al.]{markov2012weighted}
Markov, NT, Ercsey-Ravasz, MM, Gomes, AR~Ribeiro, Lamy, C, Magrou, L, Vezoli,
  J, Misery, P, Falchier, A, Quilodran, R, Gariel, MA, et~al.
\newblock A weighted and directed interareal connectivity matrix for macaque
  cerebral cortex.
\newblock \emph{Cerebral Cortex}, pp.\  bhs270, 2012.

\bibitem[Newman(2010)]{newman2010networks}
Newman, Mark.
\newblock \emph{Networks: an introduction}.
\newblock Oxford University Press, 2010.

\bibitem[Nishimoto et~al.(2011)Nishimoto, Vu, Naselaris, Benjamini, Yu, and
  Gallant]{nishimoto2011reconstructing}
Nishimoto, Shinji, Vu, An~T, Naselaris, Thomas, Benjamini, Yuval, Yu, Bin, and
  Gallant, Jack~L.
\newblock Reconstructing visual experiences from brain activity evoked by
  natural movies.
\newblock \emph{Current Biology}, 21\penalty0 (19):\penalty0 1641--1646, 2011.

\bibitem[Park \& Friston(2013)Park and Friston]{park2013structural}
Park, Hae-Jeong and Friston, Karl.
\newblock Structural and functional brain networks: from connections to
  cognition.
\newblock \emph{Science}, 342\penalty0 (6158):\penalty0 1238411, 2013.

\bibitem[Roebroeck et~al.(2005)Roebroeck, Formisano, and
  Goebel]{roebroeck2005mapping}
Roebroeck, Alard, Formisano, Elia, and Goebel, Rainer.
\newblock Mapping directed influence over the brain using granger causality and
  fmri.
\newblock \emph{Neuroimage}, 25\penalty0 (1):\penalty0 230--242, 2005.

\bibitem[Ryali et~al.(2013)Ryali, Chen, Supekar, and
  Menon]{ryali2013parcellation}
Ryali, Srikanth, Chen, Tianwen, Supekar, Kaustubh, and Menon, Vinod.
\newblock A parcellation scheme based on von mises-fisher distributions and
  markov random fields for segmenting brain regions using resting-state fmri.
\newblock \emph{Neuroimage}, 65:\penalty0 83--96, 2013.

\bibitem[Salvador et~al.(2005)Salvador, Suckling, Coleman, Pickard, Menon, and
  Bullmore]{salvador2005neurophysiological}
Salvador, Raymond, Suckling, John, Coleman, Martin~R, Pickard, John~D, Menon,
  David, and Bullmore, ED.
\newblock Neurophysiological architecture of functional magnetic resonance
  images of human brain.
\newblock \emph{Cerebral cortex}, 15\penalty0 (9):\penalty0 1332--1342, 2005.

\bibitem[Schl{\"o}sser et~al.(2003)Schl{\"o}sser, Gesierich, Kaufmann,
  Vucurevic, Hunsche, Gawehn, and Stoeter]{schlosser2003altered}
Schl{\"o}sser, Ralf, Gesierich, Thomas, Kaufmann, Bettina, Vucurevic, Goran,
  Hunsche, Stefan, Gawehn, Joachim, and Stoeter, Peter.
\newblock Altered effective connectivity during working memory performance in
  schizophrenia: a study with fmri and structural equation modeling.
\newblock \emph{Neuroimage}, 19\penalty0 (3):\penalty0 751--763, 2003.

\bibitem[Seeley et~al.(2007)Seeley, Menon, Schatzberg, Keller, Glover, Kenna,
  Reiss, and Greicius]{seeley2007dissociable}
Seeley, William~W, Menon, Vinod, Schatzberg, Alan~F, Keller, Jennifer, Glover,
  Gary~H, Kenna, Heather, Reiss, Allan~L, and Greicius, Michael~D.
\newblock Dissociable intrinsic connectivity networks for salience processing
  and executive control.
\newblock \emph{The Journal of neuroscience}, 27\penalty0 (9):\penalty0
  2349--2356, 2007.

\bibitem[Smith et~al.(2011)Smith, Miller, Salimi-Khorshidi, Webster, Beckmann,
  Nichols, Ramsey, and Woolrich]{smith2011network}
Smith, Stephen~M, Miller, Karla~L, Salimi-Khorshidi, Gholamreza, Webster,
  Matthew, Beckmann, Christian~F, Nichols, Thomas~E, Ramsey, Joseph~D, and
  Woolrich, Mark~W.
\newblock Network modelling methods for fmri.
\newblock \emph{Neuroimage}, 54\penalty0 (2):\penalty0 875--891, 2011.

\bibitem[Stone et~al.(2002)Stone, Porrill, Porter, and
  Wilkinson]{stone2002spatiotemporal}
Stone, JV, Porrill, J, Porter, NR, and Wilkinson, ID.
\newblock Spatiotemporal independent component analysis of event-related fmri
  data using skewed probability density functions.
\newblock \emph{NeuroImage}, 15\penalty0 (2):\penalty0 407--421, 2002.

\bibitem[Tomasi \& Volkow(2011)Tomasi and Volkow]{tomasi2011functional}
Tomasi, Dardo and Volkow, Nora~D.
\newblock Functional connectivity hubs in the human brain.
\newblock \emph{Neuroimage}, 57\penalty0 (3):\penalty0 908--917, 2011.

\bibitem[V{\'e}rtes et~al.(2012)V{\'e}rtes, Alexander-Bloch, Gogtay, Giedd,
  Rapoport, and Bullmore]{vertes2012simple}
V{\'e}rtes, Petra~E, Alexander-Bloch, Aaron~F, Gogtay, Nitin, Giedd, Jay~N,
  Rapoport, Judith~L, and Bullmore, Edward~T.
\newblock Simple models of human brain functional networks.
\newblock \emph{Proceedings of the National Academy of Sciences}, 109\penalty0
  (15):\penalty0 5868--5873, 2012.

\bibitem[Vincent et~al.(2007)Vincent, Patel, Fox, Snyder, Baker, Van~Essen,
  Zempel, Snyder, Corbetta, and Raichle]{vincent2007intrinsic}
Vincent, JL, Patel, GH, Fox, MD, Snyder, AZ, Baker, JT, Van~Essen, DC, Zempel,
  JM, Snyder, LH, Corbetta, M, and Raichle, ME.
\newblock Intrinsic functional architecture in the anaesthetized monkey brain.
\newblock \emph{Nature}, 447\penalty0 (7140):\penalty0 83--86, 2007.

\bibitem[Vincent et~al.(2006)Vincent, Snyder, Fox, Shannon, Andrews, Raichle,
  and Buckner]{vincent2006coherent}
Vincent, Justin~L, Snyder, Abraham~Z, Fox, Michael~D, Shannon, Benjamin~J,
  Andrews, Jessica~R, Raichle, Marcus~E, and Buckner, Randy~L.
\newblock Coherent spontaneous activity identifies a hippocampal-parietal
  memory network.
\newblock \emph{Journal of neurophysiology}, 96\penalty0 (6):\penalty0
  3517--3531, 2006.

\bibitem[Vincent et~al.(2008)Vincent, Kahn, Snyder, Raichle, and
  Buckner]{vincent2008evidence}
Vincent, Justin~L, Kahn, Itamar, Snyder, Abraham~Z, Raichle, Marcus~E, and
  Buckner, Randy~L.
\newblock Evidence for a frontoparietal control system revealed by intrinsic
  functional connectivity.
\newblock \emph{Journal of neurophysiology}, 100\penalty0 (6):\penalty0
  3328--3342, 2008.

\bibitem[Zhang et~al.(2001)Zhang, Brady, and Smith]{zhang2001segmentation}
Zhang, Yongyue, Brady, Michael, and Smith, Stephen.
\newblock Segmentation of brain mr images through a hidden markov random field
  model and the expectation-maximization algorithm.
\newblock \emph{Medical Imaging, IEEE Transactions on}, 20\penalty0
  (1):\penalty0 45--57, 2001.

\end{thebibliography}
\bibliographystyle{iclr2015}

\end{document}